\documentclass[conference]{IEEEtran}

\usepackage{cite}

\ifCLASSINFOpdf
\usepackage[pdftex]{graphicx}
\else
\fi
\usepackage[cmex10]{amsmath}
\usepackage{amsfonts}

\usepackage{fancyhdr} 
\begin{document}
%
\title{Bootstrapping Intrinsically Motivated Learning \\with Human Demonstration}

\author{\IEEEauthorblockN{Sao Mai Nguyen, Adrien Baranes and Pierre-Yves Oudeyer }
\IEEEauthorblockA{Flowers Team, INRIA  Bordeaux - Sud-Ouest, France}
}

%


\maketitle

\thispagestyle{empty}
\thispagestyle{fancy}
\lhead{}
\chead{\vspace{-40pt}
\texttt{\scriptsize{S. M. Nguyen, A. Baranes, P.-Y. Oudeyer (2011), Bootstrapping Intrinsically Motivated Learning with Human Demonstrations, in proceedings of the IEEE International Conference on Development and Learning.
 }}
\vspace{5pt}}
\rhead{}
\cfoot{}

\begin{abstract}
This paper studies the coupling of internally guided learning and social interaction, and more specifically the improvement owing to demonstrations of the learning by intrinsic motivation. We present Socially Guided Intrinsic Motivation by Demonstration (SGIM-D), an algorithm for learning in continuous, unbounded and non-preset environments. After introducing social learning and intrinsic motivation, we describe the design of our algorithm, before showing through a fishing experiment that SGIM-D efficiently combines the advantages of social learning and intrinsic motivation to gain a wide repertoire while being specialised in specific subspaces. 

\end{abstract}


%
\IEEEpeerreviewmaketitle

\section{Approaches for Adaptive Personal Robots}
The promise of personal robots operating in human environments to interact with people on a daily basis points out the importance of adaptivity of the machine to its environment and users. The robot can no longer simply be all-programmed in advance by engineers, and reproduce only actions  predesigned in factories. It needs to match its behaviour and learn new skills as the environment and users' needs change.

In order to learn an open-ended repertoire of skills, developmental robots, like animal or human infants, need to be endowed with task-independent mechanisms which push them to explore new activities and new situations \cite{Weng01, Asada09}. 
The set of skills that could be learnt is actually infinite, and can not be completely learnt within a life-time. Thus, deciding how to explore and what to learn becomes crucial. Exploration strategies, mechanisms and constraints in recent years can be classified into two broad interacting families: 1) socially guided exploration; 2) internally guided exploration and in particular intrinsically motivated exploration.

\subsection{Socially Guided Exploration}
In order to build a robot that can learn and adapt to human environment, the most straightforward way is probably to transfer knowledge about tasks or skills from a human into a machine.  That is why several  works  incorporate human input to a machine learning process. Many prior systems are strongly dependent on human guidance, unable to learn in the absence of human interaction, such as in some examples of learning by demonstration \cite{ChernovaVelosoJAIR09, Lopes09, Cederborg10, PbDCalinon} or learning by physical guidance  \cite{Calinon07}. In such systems, the learner scarcely explores on his own to learn tasks or skills beyond what it has observed with a human. Many prior works have given a human trainer control of  the reinforcement learning reward \cite{Blumberg:2002:ILI:566654.566597,Robotic-clicker-training}, provide advice\cite{A-teaching-method-for-reinforcement-leClouse-J.-and-Utgoff-P.-arning}, or  tele-operate the agent during training \cite{Effective-reinforcement-learning-for-mobile-robots}.
However, the more dependent on the human the system, the more challenging learning from interactions with a human is, due to limitations like human patience, ambiguous human input, correspondence problems \cite{Imitation-and-Social-Learning-in-Robots-Humans-and-Animals:} etc. Increasing the learner’s autonomy from human guidance could address these limitations. This is the case of internally guided exploration methods.


\subsection{Intrinsically Motivated Exploration}
Intrinsic motivation, a particular example of internal mechanism for guiding exploration, has drawn a lot of attention recently, especially for open-ended cumulative learning of skills \cite{Weng01, Oudeyer10b}. The word \textit{intrinsic motivation} was first used in psychology to describe the capability of humans to be attracted toward different activities for the pleasure that they experience intrinsically. These mechanisms have been shown crucial for humans to autonomously learn and discover new capabilities \cite{Ryan00, Deci85, Oudeyer08}. This inspired  the creation of  fully autonomous robots \cite{Barto04,Oudeyer07,Baranes09a,Schmidhuber10,Schembri07c,Schmidhuber91} with meta-exploration mechanisms monitoring the evolution of learning performances of the robot, in order to maximise informational gain, and with heuristics defining the notion of interest \cite{Fedorov72,Cohn96,Roy01}. 


While driving an efficient progressive learning in numerous cases, most intrinsic motivation approaches address only partially the challenge of unlearnability and unboundedness \cite{OudeyerIMCleverBook}. Despite efforts in the case of continuous sensorimotor spaces, computing meaningful measures of interest still requires a sampling density which decreases the efficiency of those approaches as dimensionality grows. Even in bounded spaces, the measures of interest can be cast into a form of a non-stationary regression problem, which might face the curse-of-dimensionality \cite{Bishop07}. Thus, without additional mechanisms, the identification of learnable zones with knowledge or competence progress becomes inefficient in high-dimensions. The second limitation relates to unboundedness. Actually, whatever the measure of interest used, if it is only based on the evaluation of performances of predictive models or of skills, it is impossible to explore/sample inside all localities in a life time. Therefore, complementary developmental mechanisms need to constrain the growth of the size and complexity of practically explorable spaces, by introducing self-limits in the unbounded world and/or drive them rapidly toward learnable subspaces, such as motor synergies, morphological computation, maturational constraints as well as social guidance.

\subsection{Combining Internally Guided Exploration and Socially Guided Exploration}
Intrinsic motivation and socially guided learning are often studied separately in developmental robotics, and even in opposition to one another in psychology and educational theory. Indeed, many forms of socially guided learning can be seen as extrinsically driven learning. Yet, in the daily life of humans, the two strongly interact, and their combination could on the contrary  push off the limitations we stated above.

Social guidance can drive a learner into new intrinsically motivating spaces or activities which it may continue to explore alone and for their own sake, but might have discovered only due to social guidance. Robots may acquire new strategies for achieving those intrinsically motivated activities by observing others or by listening to their advice.
Studies in robot learning by imitation and demonstration have already developed statistical inference mechanisms allowing the inference of new task constraints \cite{Calinon07, Lopes09, Cederborg10}. These techniques could be reused by intrinsically motivated learning architectures to efficiently expand the explored spaces.

Inversely, as learning that depends highly on the teacher quickly shows limitations and would discourage the user from teaching to the robot, a need for autonomous exploration is needed.  Integrating self-exploration to social learning methods could relieve the user from overly time-consuming teaching. 
For example, while self-exploration tends to result in a broader task repertoire of skills, guided-exploration with a human teacher tends to be more specialised, resulting in fewer tasks that are learnt faster. Combining both can thus bring out a system that acquires a wide range of knowledge which is necessary to scaffold future learning with a human teacher on specifically needed tasks.

Initial work in this direction \cite{ThomazBreazeal-ConnSci08} and \cite{ThomazPhDThesis} proposes a symbolic representation of actions and environment for active learning, and stresses the importance of social dialogue through both the study of the human behaviour and transparency of the robot.  The Socially Guided Exploration's motivational drives, and social scaffolding from a human partner, bias behaviour to create learning opportunities for a hierarchical Reinforcement Learning mechanism.  However, in this work, the representation of the continuous environment by the robot is discrete and the set up is a limited and preset world, with few primitive actions possible.

We would like to address the learning in the case of  an unbounded, non-preset and continuous environment. 

This paper introduces an algorithm to deal with such spaces, by merging socially guided exploration and intrinsic motivation, called Socially Guided Intrinsic Motivation (\textbf{SGIM}).The next section describes SGIM's intrinsic motivation part before its social interaction part. Then, we present the fishing experiment and its results.

\section{Intrinsic Motivations : \\ the SAGG-RIAC Algorithm}
In this section we introduce Self-Adaptive Goal Generation-Robust Intelligent Adaptive Curiosity, an implementation of competence-based intrinsic motivations  \cite{BaranesICDL09}. 
We chose this algorithm as the intrinsic motivation part of SGIM for its efficiency in learning a wide range of skills in high-dimensional space including both easy and unlearnable subparts.  Moreover, its goal directedness allows bidirectional merging with socially guided methods based on feedback on either goal and/or means. Its ability to detect unreachable spaces also makes it suitable for unbounded spaces.

\subsection{Formalisation of the Problem}
\label{formalisation}
Let us consider a robotic system whose configurations/states are described in both a state space $X$, and an operational/task space $Y$. For given configurations $(x_1, y_1) \in X \times Y$, an action $a \in A$ allows a transition towards the new states $(x_2,y_2) \in X \times Y$.   We define the action $a$ as  a parameterised dynamic motor primitive.  While in classical reinforcement learning problems, $a $ is usually defined as a sequence of micro-actions $ a= \{a_1, a_2, ..., a_n\}$, parameterised motor primitives consist of complex closed-loop dynamical policies which are actually temporally extended macro-actions, that include at the low-level long sequences of micro-actions, but have the advantage of being controlled at the high-level only through the setting of a few parameters. The association $M: (x_1, y_1, a) \mapsto (x_2,y_2)$ corresponds to a learning exemplar that will be memorised, and the goal of our system is to learn both the forward and inverse models  of the mapping $M $.
We can also describe the learning in terms of tasks, and consider $y_2$  as a \textit {goal} which the system reaches through the \textit{means} $a$ in a given \textit {context} $(x_1,y_1)$. In the following, both descriptions will be used interchangeably.

\subsection{Global Architecture of SAGG-RIAC}
The SAGG-RIAC architecture is separated in two levels:
\begin{itemize}
\item A higher level of active learning  which decides what to learn, sets a goal $y_g $ depending on the level of achievement of previous goals, and learns at a longer time scale.
\item A lower level of active learning that attempts to reach the goals set by the higher level and learns at a shorter time scale.

\end{itemize}


\subsection{Lower Time Scale: \\ Active Goal Directed Exploration and Learning}

The \textit{Active Goal Directed Exploration and Learning} mechanism guides the system toward the goal, while:

\begin{itemize}
\item A model (inverse and/or forward) is computed during exploration and is available for later goals.
\item The selection of new actions depends on local measures of  the quality of the learnt model.
\end{itemize}

\subsection{Higher Time Scale: \\ Goal Self-Generation and Self-Selection}
The Goal Self-Generation and Self-Selection process relies on feedback defined by the competence, and more precisely on the competence improvement in given subspaces of $Y$. 

\subsubsection{Competence for a Reaching Attempt}
Let  $Sim$ represent the similarity between the  final state $y_2$ of  the reaching attempt, and the actual goal $y_g$; let us note $\rho$ the other constraints. Its exact definition depends on the specific problem, but $Sim$ is to be defined in $[-\infty; 0]$, such that  the higher $Sim(y_g,y_f, \rho)$, the more efficient the reaching attempt is. 

We define the measure of competence $\gamma_{y_g}$ with respect to $Sim(y_g,y_f, \rho)$:

\vspace{-0.4cm}

\begin{eqnarray}
\gamma_{y_g}= \left\{
\begin{array}{ll}
 Sim(y_g,y_f, \rho) & \mbox{if} \ Sim(y_g,y_f, \rho) \le \varepsilon_{sim} < 0\\
  0   & \mbox{otherwise} 
 \end{array}
 \right.
 \label{competence}
\end{eqnarray}

\vspace{-0.2cm}

where $\varepsilon_{sim}$ is a tolerance factor so that we consider that the goal is reached when $Sim(y_g,y_f, \rho) > \varepsilon_{sim}$. A high value of $\gamma_{y_g}$ (i.e. close to $0$) represents a system that is competent to reach the goal $y_g$ while respecting constraints $\rho$.

\subsubsection{Definition of Interest}



Let us consider a partition $\biguplus_i {R}_i = Y$. 
Each ${R}_i$ contains attempted goals $\{y_{t_1}, y_{t_2}, ..., y_{t_k}\}_{{R}_i}$ of competences $\{\gamma_{y_{t_1}}, \gamma_{y_{t_2}}, ..., \gamma_{y_{t_k}}\}_{{R}_i}$, indexed by their relative time order of experimentation $t_1< t_2< ...< t_k  $ inside subspace ${R}_i$. 

An estimation of interest is computed for each region $R_i$ as \textit{ the local competence progress, over a sliding time window of the $\mathbf{\zeta}$ more recent goals attempted inside ${R}_i$}:

\begin{center}
\vspace{-0.4cm}
\scriptsize
\begin{eqnarray}
interest_i  =  \frac{ \left| \left(\displaystyle \sum_{j=| {R}_i|-\zeta}^{|{R}_i|-\frac{\zeta}{2}} \gamma_{y_j} \right) - \left(\displaystyle \sum_{j=|{R}_i|-\frac{\zeta}{2}}^{|{R}_i|} \gamma_{y_j} \right) \right|}{\zeta} 
\label{interest}
\end{eqnarray}


\end{center}

\subsubsection{Goal Self-Generation Using the Measure of Interest}

The goal self-generation and self-selection mechanism carries out two different processes:
\begin{enumerate}
\item Splitting $Y$ into subspaces, so as to maximally discriminate areas according to their levels of interest.
\item Selecting the region where future goals will be chosen.
\end{enumerate}
We use a recursive split of the space, each split occurring once a maximal number of goals have been attempted inside. Each split maximizes the difference of the \textit{interest} measure in the two resulting subspaces, and easily separates areas of different interest, and thus, of different reaching difficulty.

Finally, goals are chosen according to a mix of :

\textbf{Mode(1)}:  A chosen random goal inside a region which is selected with a probability proportional to its interest value: 
\vspace{-0.4cm}

\begin{eqnarray}
P_n = \frac{ interest_n - \textbf{min}(interest_i)}{\sum_{i=1}^{|R_n|}interest_i - \textbf{min}(interest_i)}
 \label{goalSelection}
\end{eqnarray}
Where $P_n$ is the selection probability of the region ${R}_n$. \\
\textbf{Mode(2)}: A selected random goal inside the whole space $Y$. \\
\textbf{Mode(3)}:  A first selected region according to the interest value (like in $mode(1)$) and then a generated new goal close to the already experimented one which received the lowest competence estimation.


 

The goal self-generation mechanism begins by exploring randomly the task space  in order to affect different values of interest to different subparts. This is why the discovery of small reachable subparts can require the fixation of an extremely important number of goals, because of the need for discrimination of these subparts among unreachable ones. In order to resolve this kind of problem, we propose to merge intrinsic motivations with the developmental paradigms of social guidance. In the following sections,  we review different kinds of social interaction modes then describe our algorithm SGIM-D (Socially Guided Intrinsic Motivation by Demonstration).

\section{Analysis of Social Interaction Modes}
Within the scope of  learning the forward and the inverse models  of the mapping $ M: (x_1, y_1, a) \mapsto (x_2,y_2)$, we would like to introduce the role of a human teacher to boost the learning of the means $a$ and goals $y_2$ in the contexts $(x_1,y_1)$. Given the model estimated by the robot $M_{R}$, and by the human teacher $M_{H}$, we can consider social interaction as a transformation $SocInter:  (M_R, M_H) \mapsto  (M2_R, M2_H) $. The goal of the learning is that the robot acquires a perfect model of the world, i.e. that  $SocInter(M_R, M_H) = (M_{perfect},M_{perfect})$. The social interaction is a combination of these behaviours:
\begin{itemize}
\item the human teacher's behaviour $SocInter_H$ in response to the visible state of the robot and the environment. 
\item the machine learner's behaviour $SocInter_R$ in response to the guidance of the human teacher.
\end{itemize}
We presume a transparent communication between the teacher and the learner, ie, the teacher can access the real visible state of the robot as a noiseless function of its internal state $visible_R(M_R)$. Let us note $\widetilde{visible}_R$ the "perfect visible state" of the robot, defined as the value of the visible states of the robot when its estimation of the model is perfect : $M_R = M_{perfect}$.
Moreover,  we simplify the general problem first by postulating that the teacher is omniscient and that his estimation of the model is the perfect model $M_{perfect}$. Therefore, our social interaction is a transformation $SocInter: M_R \mapsto M $.

In order to define the social interaction that we wish to consider, we need to examine the different possibilities.

\subsection{ Role of the Teacher}
First of all, let us define which type of interaction takes place, and what role we give to the teacher:  

\subsubsection{ The teacher provides high-level evaluation, feedback, or labels to a machine learner} : the teacher would guide the robot through an estimation of distance between the robot's visible state and its "perfect visible state" : $SocInter_H  \sim dist(visible_R,\widetilde{visible}_R) $. \cite{ThomazBreazeal-ConnSci08} used such feedback to boost reinforcement learning. Child development psychology would illustrate the importance of such feedback from teachers to infants for instance by the means of motherese  \cite{springerlink:10.1023/A:1013215010749}. Nevertheless, as in parent-child interaction cheering is completed by games where the parents show and instruct children interesting cases, and help children reach their goal, a more informational interaction would better help the learner than mere cheering.

\subsubsection{ The teacher shows how to reach the goal that the robot aims at} the teacher here would show to the robot a means to reach the goal that the robot had set by itself: $SocInter_H(x_1,y_1,y_2) \in \{ a| \exists x_2 : l(x_1, y_1, a) = (x_2,y_2) \}$. An applicable case is the example of active learning where the robot asks for demonstrations  \cite{ChernovaVelosoJAIR09} when it makes no progress and does not reach the goal it has set by itself. The robot learns new ways to reach that goal and can replicate the action. This is an imitation behaviour in a restricted definition of the term, where the observer copies the specific motor patterns.

\subsubsection{ The teacher shows a context  (new initial conditions)}  $SocInter_H = (x_1,y_1) \in X \times Y$ . The teacher here could set up new situations and contexts, and let the robot learn autonomously in the demonstrated context. This setting would be interesting for a mobile robot that changes location such as exploration, rescue or space robots.

\subsubsection{The teacher demonstrates goals} such as in \cite{Cederborg10}, i.e. $SocInter_H = y_2 \in Y $. This would typically help a robot that has been trying to solve tasks of low interest values (the measure of the level of interest depends on the specific experiment). It learns about results and changes that can be accomplished in the environment and attempts to replicate such states and changes. This is the definition of an emulation behaviour, one of the two broad categories of social learning along with imitation \cite{Imitation-in-animals-and-artifacts-chapter-Three-sources-of-information-in-social-learning}. Nevertheless, emulation alone can not satisfyingly represent social learning, as young children are prone to imitate the action sequences, even parts that are not obviously necessary to achieve the goal: a phenomenon known as over-imitation \cite{springerlink:10.1007/s10071-004-0239-6}. 

\subsubsection{The teacher shows both a means and a goal} $SocInter_H \in A \times Y $. This is a typical imitation behaviour in the broad sense, where the observer copies both the specific motor patterns and consequent results that are jointly inferred to have been part of the behaviour intention. The new sample  highlights  a subspace  which the robot can explore. This seems to be the most complete approach as it enables both imitation and emulation, as it influences the learner both from the action point of view and the goal point of view.

To sum up, the teacher who shows both a means and a goal seems to offer the best opportunity for the learner to progress, for he provides the learner with both example goals and example means, so that the learner can use both the means and/or the goal-driven approach.

\subsection{Timing of the Social Interaction}
After these considerations about the nature human teacher's behaviour and guidance $SocInter_H$, our next question is: when should the interaction take place?
 
\subsubsection{ In the very beginning} before any personal experience of the robot itself. This would speed up the learning from the beginning, but has no merit as it would not account for the adaptability and flexibility to the changing environment and demand from the user.
\subsubsection{ At a regular pace} (every N experiments). This would represent the regular and continuous social interactions the system has with its teacher, and is best to assess quantitatively the improvement of its learning.
\subsubsection{ When the robot stops making progress} the measure of progress being specific to the learning problem. Either it asks for help by himself (sends a non null $SocInter_R$ ), or the benevolent teacher steps in. This seems the best solution to maximise the utility of the teacher, but brings questions such as how to evaluate that the robot is stuck, and at which level of difficulty the teacher should step in. It would also assume that the teacher is attentive to the state of the robot.

Although the 3rd case seems interesting theoretically, as the purpose of this work is to compare the performance of different algorithms, we opted for an idealised teacher, who would have continuous interaction with the robot throughout the learning duration. And to make the teaching neutral and not biased to fit our algorithm specifically, we choose non optimal teaching parameters. The teacher gives a demonstration at constant frequency, and randomly selects it from a set of demonstrations.
 
\subsection{Which Demonstrations to Choose?}
This brings us to the more specific question of which demonstrations among all the possible demonstrations, the teacher should give to the learner:

\subsubsection { One sample among a set of completely random examples} this seems the easiest solution but the teaching would not differ from random exploration.
\subsubsection { One random among the unreached goals} this solution makes the robot explore new goals and unexplored subspaces.
\subsubsection{  The farthest among the unreached goals } it would make sure the new goal provided is not already accessible to the robot, but still, it would prove to be too difficult a goal to help the robot progress.
\subsubsection{  The nearest among the unreached goals}  it respects the progressive development idea, but demonstrations would fail to introduce the learner to new unexplored subspaces.

To bootstrap a system endowed with intrinsic motivation, we choose to use a learning by demonstration of means and goals, where the teacher introduces at regular pace a random demonstration among the unreached goals.

\section{SGIM Algorithm}

\begin{figure}
\centering
\includegraphics[width=0.5\textwidth]{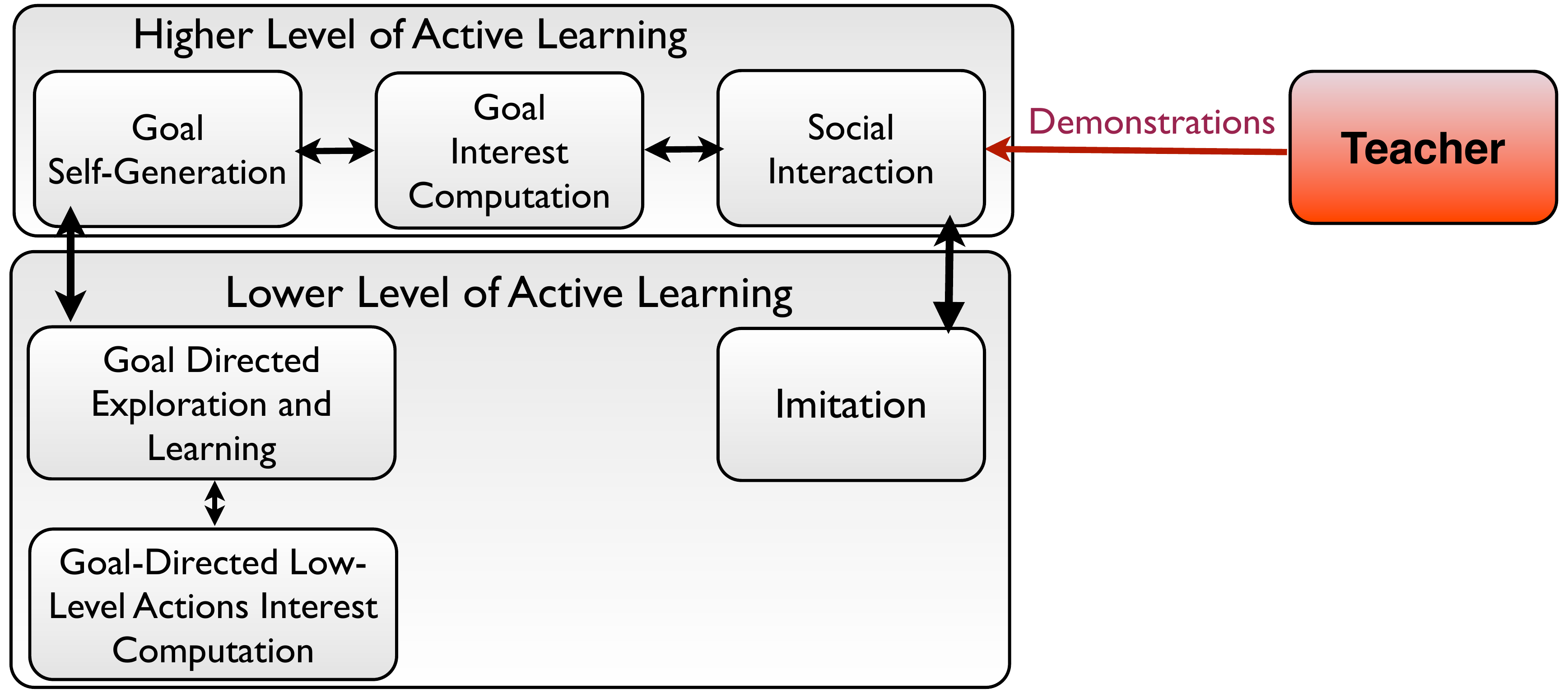}
\caption{Structure of SGIM-D (Socially Guided Intrinsic Motivation by Demonstration). SGIM-D is organised into 2 levels.}
\label{StructureSGIM}
\vspace{-0.6cm}
\end{figure}

This section details SGIM as an algorithm for the learning of an inverse model in a continuous, unbounded and non-preset framework, combining both intrinsic motivation and  social interaction. Our Socially Guided Intrinsic Motivation Algorithm merges the SAGG-RIAC algorithm of intrinsic motivation with a learning by demonstration as social interaction.
The system includes two different levels of learning (fig. \ref{StructureSGIM}).

\subsection{Higher level of Learning}
The higher level of active learning  decides which goal $(x_2,y_2)$ is interesting to explore. It contains 3 modules. The \textit{Goal Self-Generation module} and the \textit {Goal Interest Computation} module are as in SAGG-RIAC. The \textit{Social Interaction} module manages the interaction with the human teacher. It interfaces between the social guidance of the human teacher $SocInter_H$ and the goal interest computation module of intrinsic motivation to decide which lower level behaviour should be triggered. With the choices of social interaction mode we choose, it interrupts the intrinsic motivation  at every demonstration by the teacher. It first triggers an emulation effect, as it registers the demonstration $(a_{demo},y_{demo})$ in the memory of the system and gives it as input to the goal interest computation module. It also triggers the imitation behaviour and sends the demonstrated action $a_{demo}$ to the imitation module.

\subsection{Lower Level of Learning}
The lower level of active learning  also contains 3 modules. The \textit {Goal Directed Exploration and Learning module} and the \textit{Goal Directed Low Level Actions Interest Computation} module are as in SAGG-RIAC.  The \textit {Imitation} module interfaces with the high-level social interaction module. It takes as input an action $a_{demo}$, and tries to repeat it a fixed number of times, with variations in order to explore the locality of $a_{demo}$.

The above description is detailed for our choice of SGIM by Demonstration. Such a structure would remain suitable for other choices of social interaction modes, and we only have to change the content of the Social Interaction module, and change the Imitation module to the chosen behaviour. Our structure, notably, can deal with cases where the intrinsically motivated part gives a feedback to the teacher, as the Goal Interest Computation module and the Social Interaction module communicate bilaterally. For instance, the case of active learning we mentioned in the analysis of social interaction modes, where the learner asks the teacher for demonstrations, can still use the structure presented.

We have until now, discussed intrinsic motivation and more specifically the SAGG-RIAC algorithm, and we have analysed social learning and its different modes to design Socially  Guided Intrinsic Motivation by Demonstration (SGIM-D) that merges both paradigms, and to learn a model in a continuous, unbounded and non-preset framework. In the following section we use SGIM-D to learn a fishing skill.

\section{Fishing Experiment}

 This  fishing experiment focuses on the learning of inverse models in a continuous space, and deals with a very high-dimensional and redundant model. The model of a fishing rod in a simulator might possibly be mathematically computed, but a real-world fishing rod's dynamics would be impossible to model. A learning system of such case is therefore interesting.

\subsection{Experimental Setup}
\begin{figure}
\centering
\includegraphics[width=0.32\textwidth]{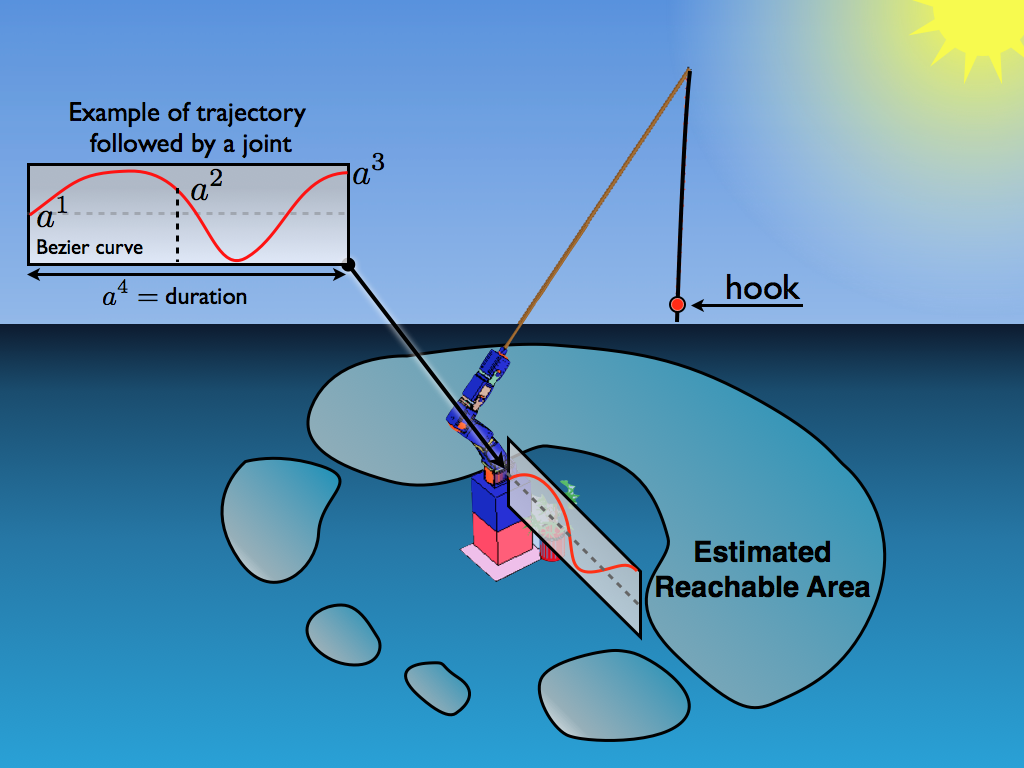}
\caption{Fishing experimental setup.}
\label{FishingRod}
\vspace{-0.6cm}
\end{figure}

Our continuous environment is a 6 degrees-of-freedom robot arm that learns to use a fishing rod (fig. \ref{FishingRod}) to know, for a given goal position $y_g$, where the hook should reach when falling into the water and which action $a$  to perform. This is an inverse model in a continuous and unbounded environment of complex system that can hardly be described by physical equations.

In our experiment, $X$ describes the actuator/joint positions and the state of the fishing rod. $ Y$ is a 2-D space that describes the position of the hook when it reaches the water. The robot always starts with the same initial position,  $x_1$ and $y_1$ always take the same values $x_{org}$ and $y_{org}$. Variable $a$ describes the parameters of the commands for the joints. In our setup, we choose to control each joint with a Bezier curve defined by 4 scalars (initial, middle and final joint position and a duration). Therefore an action is represented by a $ 6 \times 4= 24$  parameters: $a= (a^1,a^2, ...a^{24} )$. Because our experiment uses for each trial the same context $(x_{org}, y_{org})$, our system memorises after executing  every action $a$, simply the context-free association $ a \mapsto y_2$ using a combination of social learning and intrinsic motivation.

The experimental scenario sets the robot to explore the task space through intrinsic motivation when it is not interrupted by the teacher. After $P$ movements, the teacher interrupts whatever the robot is doing, and gives him an example $(a_{demo}, y_{demo})$. The robot first registers that example in its memory as if it were its own. Then, the Imitation module tries to imitate the teacher with movement parameters  $a_{imitate} = a_{demo} + a_{rand} $ with $a_{rand}$ a random movement parameter variation, so that $|a_{rand}|< \epsilon$. At the end of the imitation phase, SGIM-D shifts back to the autonomous exploration mode which is based on a measure of competence, specific to the problem and that we define hereafter.

\subsection{Measure of Competence}
\label{one-goal}
Let us first consider that the robot learns to reach a fixed goal position $y_g = (y_g^1,y_g^2)$. We define the similarity function $Sim$ and thus the competence as linked with the euclidian distance  between the final state and the goal in the task space after a reaching attempt $D(y_g,y_2)$, and normalised by the distance between the origin position $y_{org}$ and the goal: $ D(y_{org}, y_g)$. This allows, for instance, to give the same competence level when considering a goal at 1km from the origin position that the robot approaches at 0.1km, and a goal at 100m that the robot approaches at 10m. 

$D(y_1,y_2)$ is the euclidian distance rescaled to [0;1]. Each dimension thus has the same weight in the estimation of competence. The similarity measure is defined as:

\vspace{-0.5cm}
\begin{eqnarray}
Sim(y_g, y_2, y_{org}) = \left\{
\begin{array}{ll}
 -1 & \mbox{if}  \frac{D(y_g,y_2)}{D(y_g, y_{org})}  > 1\\
   - \frac{D(y_g,y_2)}{D(y_g, y_{org})}  & \mbox{otherwise}  
 \end{array}
 \right.
\end{eqnarray}
 \label{similarity}
\vspace{-0.4cm}

Reaching a goal $y_g$ requires movement parameters $a$ leading to this chosen state $y_g$. 
Here, our direct model $M :a \mapsto y$ only considers the 24 parameters $a= (a^1,a^2, ...a^{24} )$ as inputs of the system, and a position in $ (y^1,y^2)$ as output. In this experiment, we wish to estimate the inverse model $InvM :  y \mapsto a $ and use the following optimisation mechanism which can be divided into two different regimes:
 
\subsubsection{Exploitation Regime}
The exploitation regime uses the memory data to interpolate an inverse model $ InvM: (y^1,y^2) \rightarrow  (a^1,a^2, ...a^{24} ) $.  Given the high redundancy of the problem, we choose a local approach and extract the potentially more reliable data using the following method. First, we compute the set $L$ of the $l_{max}$ nearest neighbours of $y_g$ and their corresponding movement parameters using an ANN method \cite{Muja09}, which is based on a tree split using the k-means process:
\vspace{-0.3cm}
\begin{eqnarray}
L &=&  \left\{  (y,a)_1,  (y,a)_2,   ... ,  (y,a)_{l_{max}}   \right\}   \subset (Y\times A)^{l_{max}} 
\end{eqnarray}
Then, for each element $(y,a)_l \in L$, we compute its reliability. Let us consider the set $K_l$ which contains the $k_{max}$ nearest neighbours of $x_l$ :
\vspace{-0.4cm}
\begin{eqnarray}
K_l &=& \left\{  (y,a)_1,  (y,a)_2,   ... ,  (y,a)_{k_{max}}   \right\}
\end{eqnarray}
As the reliability of a movement depends both on the local knowledge of the locality and the reproductivity of it, we define it as the variance $var_l$ of the set $K_l$. We compute for each element $(y,a)_l \in L$, its reliability as $dist(y_l,y_g) + \alpha\times var_l$, where $\alpha $ is a constant set to 0.5 in our experiment. We choose the smallest value, as the most reliable set $(y,a)_{best}$.

In the locality of the set $(y,a)_{best}$,  we interpolate using the $k_{max}$ elements of $K_{best}$ to compute the action corresponding to $y_g$ :
$a_g  = \sum_{k=1}^{k_{max}} {coef_k a_k} $ where  $coef_k \sim Gaussian(dist(y_k,y_g)) $ is a normalized gaussian of the euclidian distance between $y_k$ and the goal $y_g$.

We execute action $a_g$ and continue with the Nelder-Mead simplex algorithm \cite{Lagarias1998SJO}, to minimise the distance of the final state $y_2$ to the goal $y_g$. This algorithm uses a simplex of n + 1 points for n-dimensional vectors x. It first makes a simplex around the initial guess $a_g$ with the $a_k, k= 1,...k_{max}$. It then updates the simplex with points around the locality until the distance to minimise is below a threshold.

\subsubsection{Exploration Regime}
In this regime the system just uses a random movement parameter to explore the space.

The system continuously estimates the distance between the goal $y_g$ and the closest already reached position $y_c$: $dist(y_c, y_g)$. The system has a probability proportional to  $dist(y_c, y_g)$ of being in the exploration regime, and the complementary probability of being in the exploitation regime.

\subsection{Simulations}
All the experimental setup has been designed for a human teacher. Nevertheless, to test our algorithm, to control better the demonstrations of the teacher and to be able to collect statistics, we start by experimenting on V-REP physical simulator, which uses a ODE physics engine that updates every 50 ms.  The noise of the control system of the 3D robot is estimated to 0.073 for  measures of 10 attempts of each of the 20 random movement parameters, while the reachable area spans between -1 and 1 for each dimension. 

\begin{figure}
\centering
\includegraphics[width=0.27\textwidth]{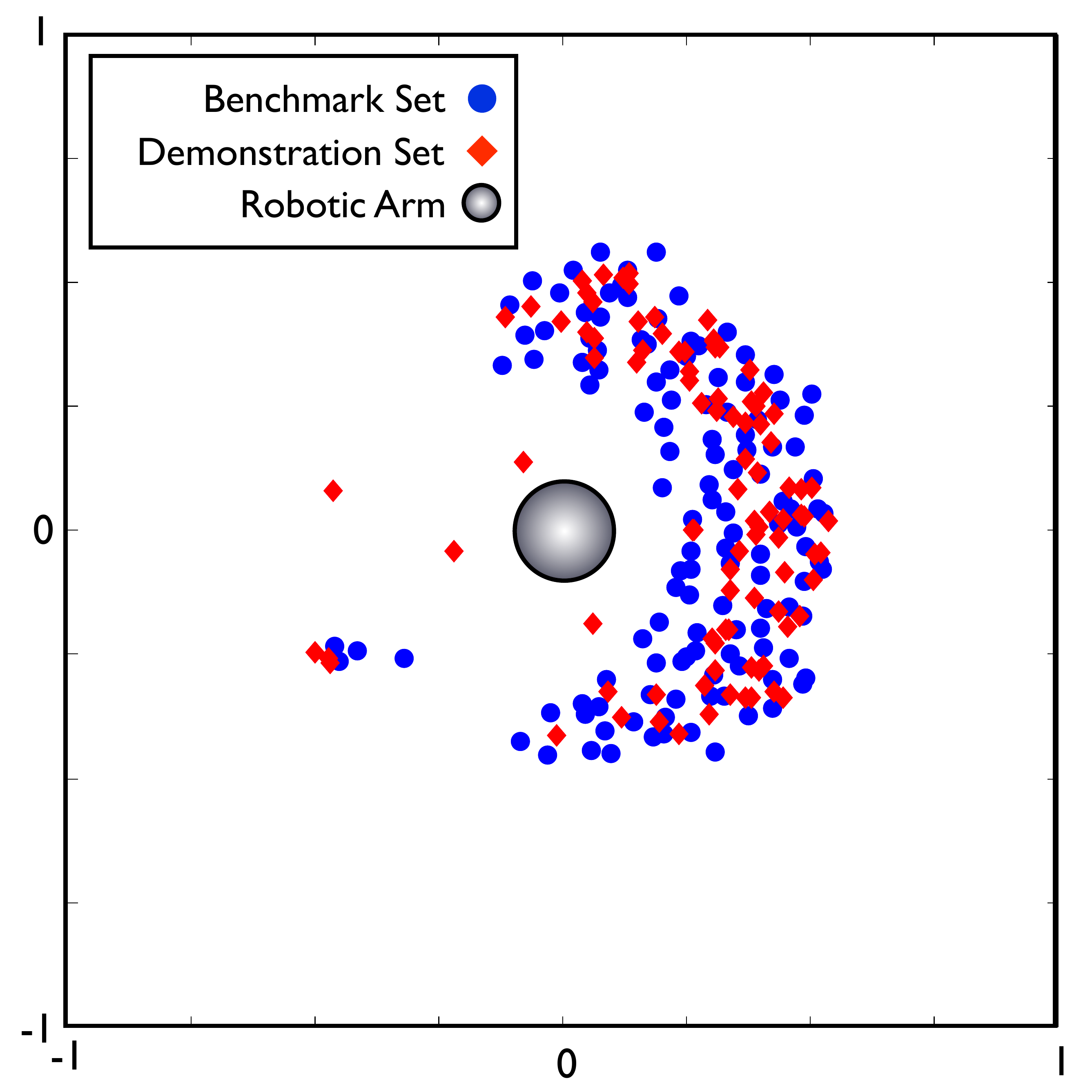}
\caption{Maps of the benchmark points used to assess the performance of the robot, and the teaching set, used in SGIM.}
\label{BenchmarkTeachingSet}
\vspace{-0.6cm}
\end{figure}

After several runs of random explorations and SAGG-RIAC, we determined the apparent reachable space as the set of all the reached points in the goal/task space, which makes up some 70 000 points. We then divided the space into small squares, and generated a point randomly in each square. Using a $26\times 16$ grid, we obtained a set of 129 goal points in the task space, representative of the reachable space, and independent of the experiment data used  (fig. \ref{BenchmarkTeachingSet}) .

Likewise, we prepared a teaching set. With the perspective that the demonstrations should be recorded on the robot via kinesthetic teaching, the robot has access to the action parameters, without having to compute the inverse kinematics.  In our simulation, we provided the robot with demonstrations that are both action parameters $a$ and goal $y$, using the data of several runs of random explorations and SAGG-RIAC. To define the 27 demonstration points (fig. \ref{BenchmarkTeachingSet}), we divided the reachable space into small squares $subY$. In each $subY$,  we choose a demonstration $(a,y), y \in subY$. So that the teacher gives the best replicable demonstration, we compute  $M_H^{-1}(subY)= \{a | M_H: a \mapsto y\in subY \}$. We tested all the movement parameters $a \in M_H^{-1}(subY)$  to choose the most reliable one, ie, that resulted in the smallest variance in the goal space $a_{demo}= min \{ var( M_H(a)) ) \}_{a \in  M_H^{-1}(subY)} $.

\subsection{Experimental results}
\begin{figure}
\centering
\includegraphics[width=0.5\textwidth]{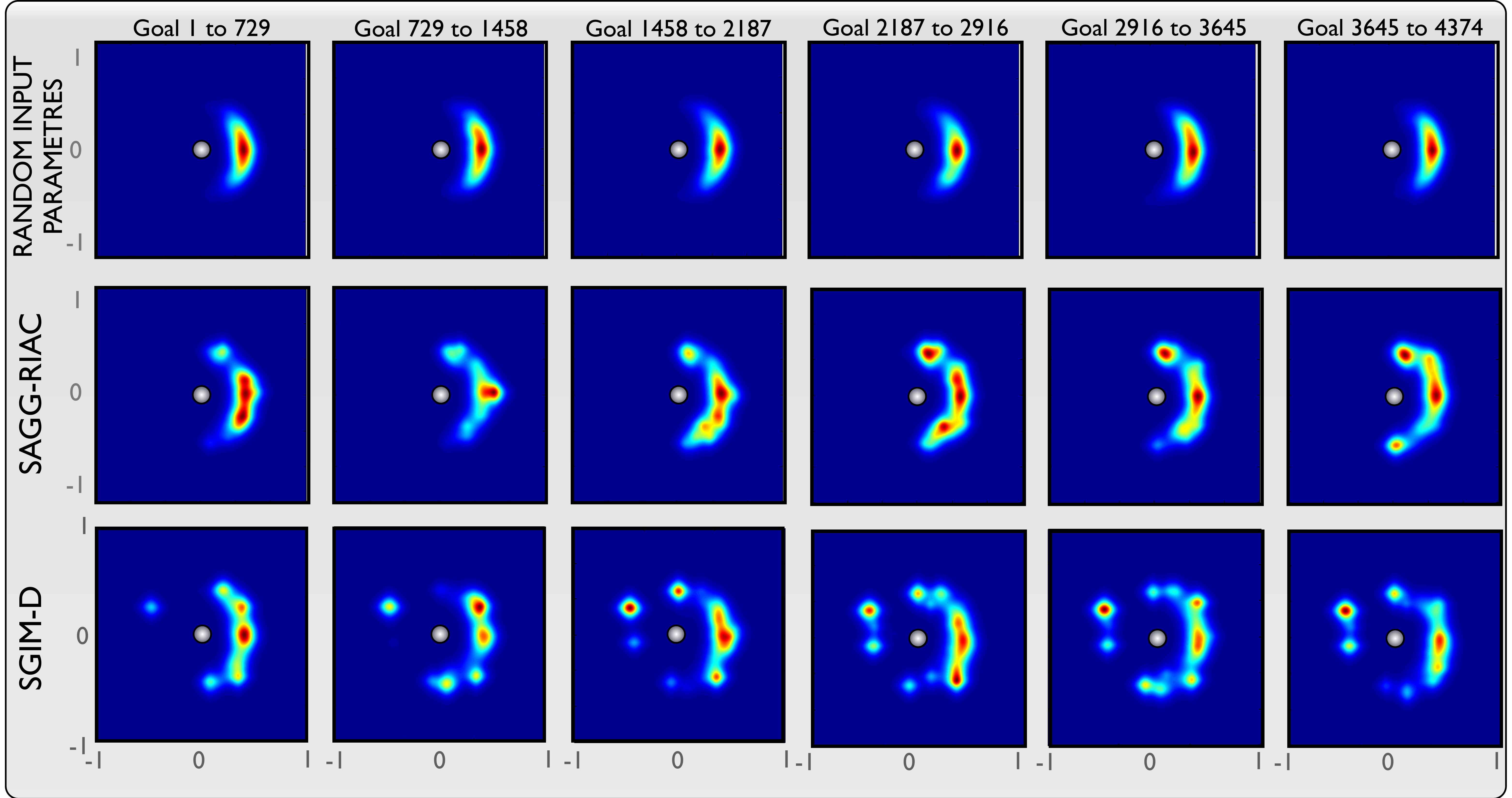}
\caption{Histograms of the positions explored by the fishing rod inside the 2D goal space $(y^1,y^2)$. Each row shows the timeline of the cumulated set of points throughout 5000 random movements. Each row represents a different learning algorithm : random input parameters, SAGG RIAC and SGIM-D. }
\label{FigureSGIM}
\vspace{-0.0cm}
\end{figure}

\begin{figure}
\centering
\includegraphics[width=0.5\textwidth]{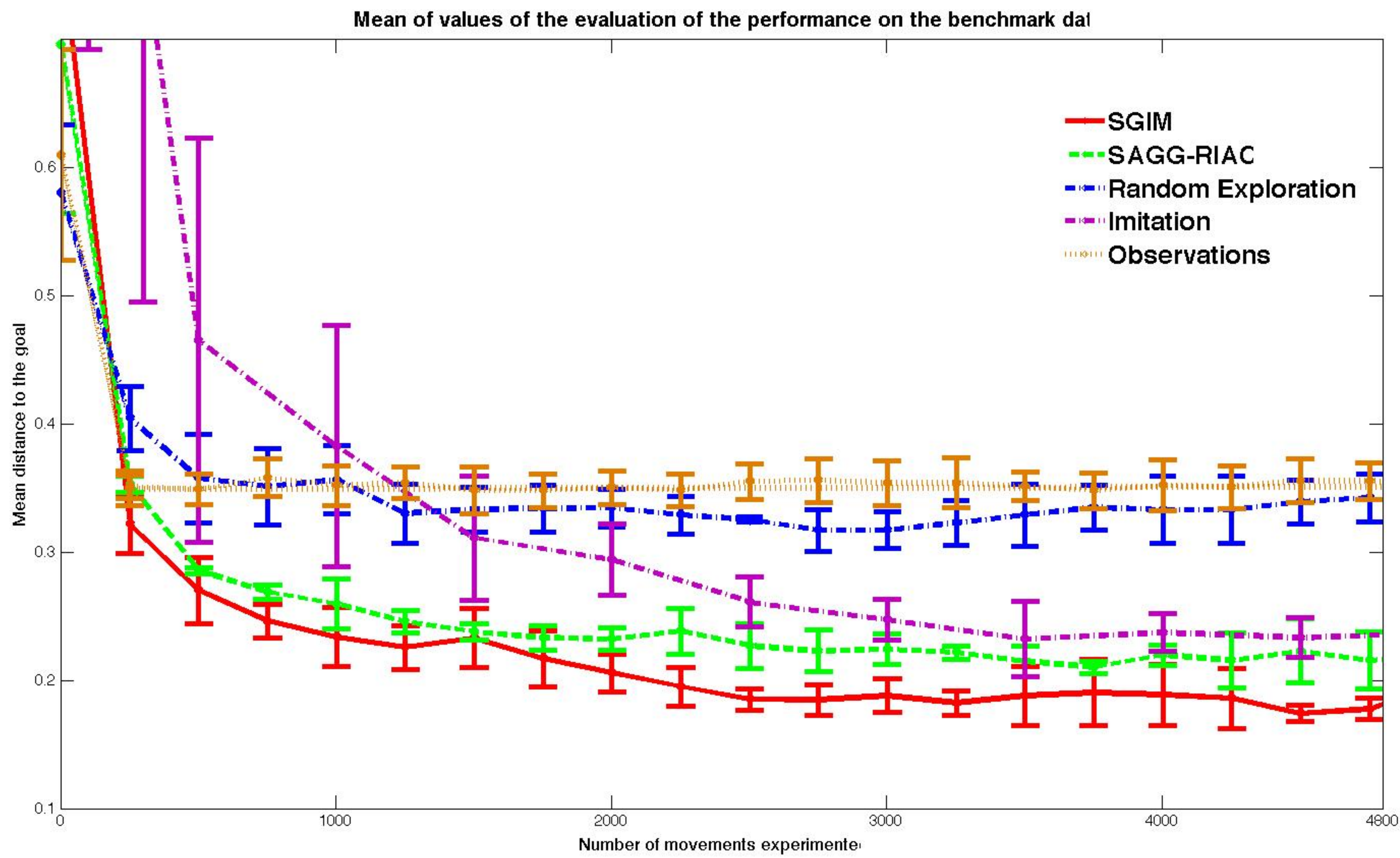}
\caption{Evaluation of the performance of the robot under the learning algorithms: demonstrations only, random exploration, SAGG-RIAC and SGIM-D. We plotted the mean distance to the benchmark points over several runs of the experiment.}
\label{CompareEvaluationInterpolation}
\vspace{0.0cm}

\end{figure}

We run several times the algorithms : 
\begin{itemize}
\item SGIM-D : one demonstration every 150 movements
\item SAGG-RIAC
\item learning by demonstrations only: the robot always makes small variations of the most recent demonstration.
\item random exploration: random movement parameters $a$. 
\end{itemize}
 For every simulation, 5000 movements are performed. The performance was assessed on the same benchmark set every 250 movements. We plot  the histogram of the positions of the hook in the task space when it reaches the water (fig. \ref{FigureSGIM}).  Each column represents a different timeframe, and each line represents a different learning algorithm. Fig.  \ref{CompareEvaluationInterpolation} plots  the mean error of the robot when it tries to reach a goal point defined by the benchmark. The values are averaged on all points in the benchmark, but also on different runs of the experiment.

\subsubsection{SAGG-RIAC compared to random exploration}
The 1st row of fig.  \ref{FigureSGIM} shows that a natural position lies around $(0.5,0)$ in the case of an exploration with random movement parameters. Most movements parameters map to a position of the hook around that central position. We can note that the distribution of the hook positions does not change through the different timeframes, as we expect.
 The second row  shows the histogram in the task space of the explored points under SAGG-RIAC algorithm. Compared to a random parameters exploration, SAGG-RIAC has increased the explored space, and most of all, covers more uniformly the explorable space. Besides, the exploration changes through time as the system finds new interesting subspaces to focus on and explore. Intrinsic motivation exploration has resulted in a wider repertoire for the robot. Furthermore, fig. \ref{CompareEvaluationInterpolation} shows that the robot performs significantly better with SAGG-RIAC, and can reach closer the points of the evaluation benchmark. Intrinsic motivation exploration increases precision over random exploration.
 
\subsubsection{Performance of SGIM}

Fig. \ref{CompareEvaluationInterpolation} shows that the performance of the SAGG-RIAC increases in the case of SGIM-D, but also that SGIM-D performs better than learning by demonstrations alone. Demonstrations given by the teacher improve the precision of the inverse model $InvM$ over the plain autonomous exploration or learning by demonstration only. However, the difference does not lie so much in the performance and precision of the robot, but mostly in the subspaces explored. Fig. \ref{FigureSGIM}  highlights a region around $(-0.5, -0.25)$ that was completely ignored by both the random exploration and SAGG-RIAC, but was well explored by SGIM-D. This isolated subspace corresponds to a very small  subspace in the parameters space, seldom explored by the random exploration or SAGG-RIAC. On the contrary, SGIM-D will highlight these subspaces thanks to the demonstrations. The teacher gives a demonstration that triggers the robot's interest and he will focus his attention on that area as long as exploration improves his competence in this subspace. We also note that the demonstrations occurred only once every 150 movements. Even a scant presence of the teacher can significantly improve the performance of the autonomous exploration.

In conclusion, SGIM-D improves the precision of the system even with little intervention from the teacher, and helps point out key subregions to be explored. The teacher successfully transfers his knowledge to the learner and bootstraps autonomous exploration.

\section{ Conclusion and Discussion}
 This paper introduces Socially  Guided Intrinsic Motivation by Demonstration, \textbf{SGIM-D}, a learning algorithm for models in a continuous, unbounded and non-preset framework, which efficiently combines social learning and intrinsic motivation. It takes advantage of the demonstrations of the teacher to explore unknown subspaces, and to discriminate interesting subspaces from uninteresting ones. It also takes advantage of the autonomous exploration of SAGG-RIAC to improve its performance and gain precision in the absence of the teacher in a wide range of tasks. It proposes a hierarchical learning with a higher level that determines which goals are interesting either through intrinsic motivation or social interaction, and a lower-level learning that endeavours to reach it.  Our simulation indicates that SGIM-D successfully combines learning by demonstration and autonomous exploration even in an experimental setup as complex as having a  continuous 24-dimension action space.

Nevertheless, in this initial validation study in simulation, we make strong suppositions about the teacher. He has the same motion generation rules as the robot, so that a movement demonstrated  by the teacher can theoretically be exactly represented and reproduced by the robot.
While the experiment has been designed for social interaction, only simulations have been conducted until now. Experiments with human demonstrations need to be realised and to address the problems of correspondence and of a biased teacher.

For future work, we would first like to realise the experiment  in a real world environment with a human teacher. We will then study further the effects of different parameters of  social interaction on the performance of the robot, for instance the effects of the frequency of the demonstrations given by the teacher. The parameters of the teaching, such as the rationales for selecting timing of the social interaction and demonstrations have not been chosen in this paper to optimise SGIM-D. A more precise study of these parameters could even show better performance of SGIM-D. More generally, exploring and evaluating systematically the other scenarios in which a human teacher can be involved, as mentioned in section III,  should be instructive. An interesting angle to study would also be the study of the switching between imitation and emulation. In our experiment, the robot imitates the teacher for a fixed amount of time, and afterwards, SGIM-D takes into account these new data only from the goal point of view, as in emulation. However a more natural and autonomous algorithm for switching between or combining these two modes could improve the efficiency of the system.

\section*{Acknowledgment}

This research was partially funded by ERC Grant EXPLORERS 240007 and ANR MACSi.



%

{\footnotesize
\bibliographystyle{IEEEtran}
\bibliography{./BIBLIOF2.bib}
}

\end{document}